\documentclass[11pt, a4paper, logo, copyright, nonumbering]{deepseek}
\usepackage[authoryear, compress, round]{natbib}
\usepackage{dblfloatfix}
\usepackage{ulem}
\usepackage{caption}
\usepackage{dramatist}
\usepackage{xspace}
\usepackage{pifont}
\usepackage{multirow}
\usepackage{tcolorbox}
\usepackage{xltabular}
\usepackage{longtable}
\usepackage{hyperref}
\usepackage{amsfonts}
\usepackage{amsmath}
\usepackage{amssymb}
\usepackage{lineno}

\usepackage[bottom]{footmisc}

\usepackage{CJKutf8}
\usepackage{subfigure}
\usepackage{setspace}

\definecolor{keywordcolor}{rgb}{0.7, 0.1, 0.1}   
\definecolor{tacticcolor}{rgb}{0.0, 0.1, 0.6}    
\definecolor{commentcolor}{rgb}{0.4, 0.4, 0.4}   
\definecolor{symbolcolor}{rgb}{0.0, 0.1, 0.6}    
\definecolor{sortcolor}{rgb}{0.1, 0.5, 0.1}      
\definecolor{attributecolor}{rgb}{0.7, 0.1, 0.1} 
\usepackage{listings}

\makeatletter
\def\@BTrule[#1]{%
  \ifx\longtable\undefined
    \let\@BTswitch\@BTnormal
  \else\ifx\hline\LT@hline
    \nobreak
    \let\@BTswitch\@BLTrule
  \else
     \let\@BTswitch\@BTnormal
  \fi\fi
  \global\@thisrulewidth=#1\relax
  \ifnum\@thisruleclass=\tw@\vskip\@aboverulesep\else
  \ifnum\@lastruleclass=\z@\vskip\@aboverulesep\else
  \ifnum\@lastruleclass=\@ne\vskip\doublerulesep\fi\fi\fi
  \@BTswitch}
\makeatother

\addto\extrasenglish{
}

 {\begin{list}{}%
         {\setlength{\leftmargin}{#1}}%
         \item[]%
 }
 {\end{list}}

\newcommand{\lastwork}{DeepSeek-Prover-V1}
\newcommand{\thiswork}{DeepSeek-Prover-V1.5}

\newcommand{\treesearch}{RMaxTS}

\reportnumber{001} 

\renewcommand{\today}{}

\title{\centering \thiswork: Harnessing Proof Assistant Feedback \\ for Reinforcement Learning and Monte-Carlo Tree Search}

\author[*]{
\footnotesize
Huajian Xin*, Z.Z. Ren*, Junxiao Song*, Zhihong Shao*, Wanjia Zhao, Haocheng Wang, Bo Liu, Liyue Zhang\newline
Xuan Lu, Qiushi Du, Wenjun Gao, Qihao Zhu, Dejian Yang, Zhibin Gou, Z.F. Wu, 
Fuli Luo, Chong Ruan

\small
DeepSeek-AI \\
\small
\url{https://github.com/deepseek-ai/DeepSeek-Prover-V1.5}
\vspace{-0.2in}
}
\correspondingauthor{Core contributors}

\renewcommand{\phi}{\varphi}

\renewcommand{\leq}{\leqslant}
\renewcommand{\geq}{\geqslant}

\renewcommand{\epsilon}{\varepsilon}
\renewcommand{\imath}{\mathrm{i}}

\newlength{\restsubwidth}
\newlength{\restsubheight}
\newlength{\restsubmoreheight}
\newcommand{\rest}[2]{%
        \settowidth{\restsubwidth}{\ensuremath{#2}}
        \settoheight{\restsubheight}{\ensuremath{{}_{#2}}}
        \ensuremath{{#1\hskip 0.5pt}_{\vrule\kern2pt\parbox[b][%
        4pt][b]{\the\restsubwidth}{%
                        \ensuremath{{}_{#2}}}}}
        }

\usepackage{thmtools}

\newcommand{\ie}{\textit{i.e.}}
\newcommand{\eg}{\textit{e.g.}}
\newcommand{\aka}{\textit{a.k.a.}}

\begin{abstract}
We introduce \thiswork, an open-source language model designed for theorem proving in Lean 4, which enhances \lastwork~by optimizing both training and inference processes. Pre-trained on DeepSeekMath-Base with specialization in formal mathematical languages, the model undergoes supervised fine-tuning using an enhanced formal theorem proving dataset derived from \lastwork. Further refinement is achieved through reinforcement learning from proof assistant feedback (RLPAF). Beyond the single-pass whole-proof generation approach of \lastwork, we propose \treesearch, a variant of Monte-Carlo tree search that employs an intrinsic-reward-driven exploration strategy to generate diverse proof paths. \thiswork~demonstrates significant improvements over \lastwork, achieving new state-of-the-art results on the test set of the high school level miniF2F benchmark (63.5\%) and the undergraduate level ProofNet benchmark (25.3\%).
\end{abstract}

\begin{document}
\begin{CJK*}{UTF8}{gbsn}

\maketitle

\begin{figure}[h]
   \centering
   \includegraphics[width=\textwidth]{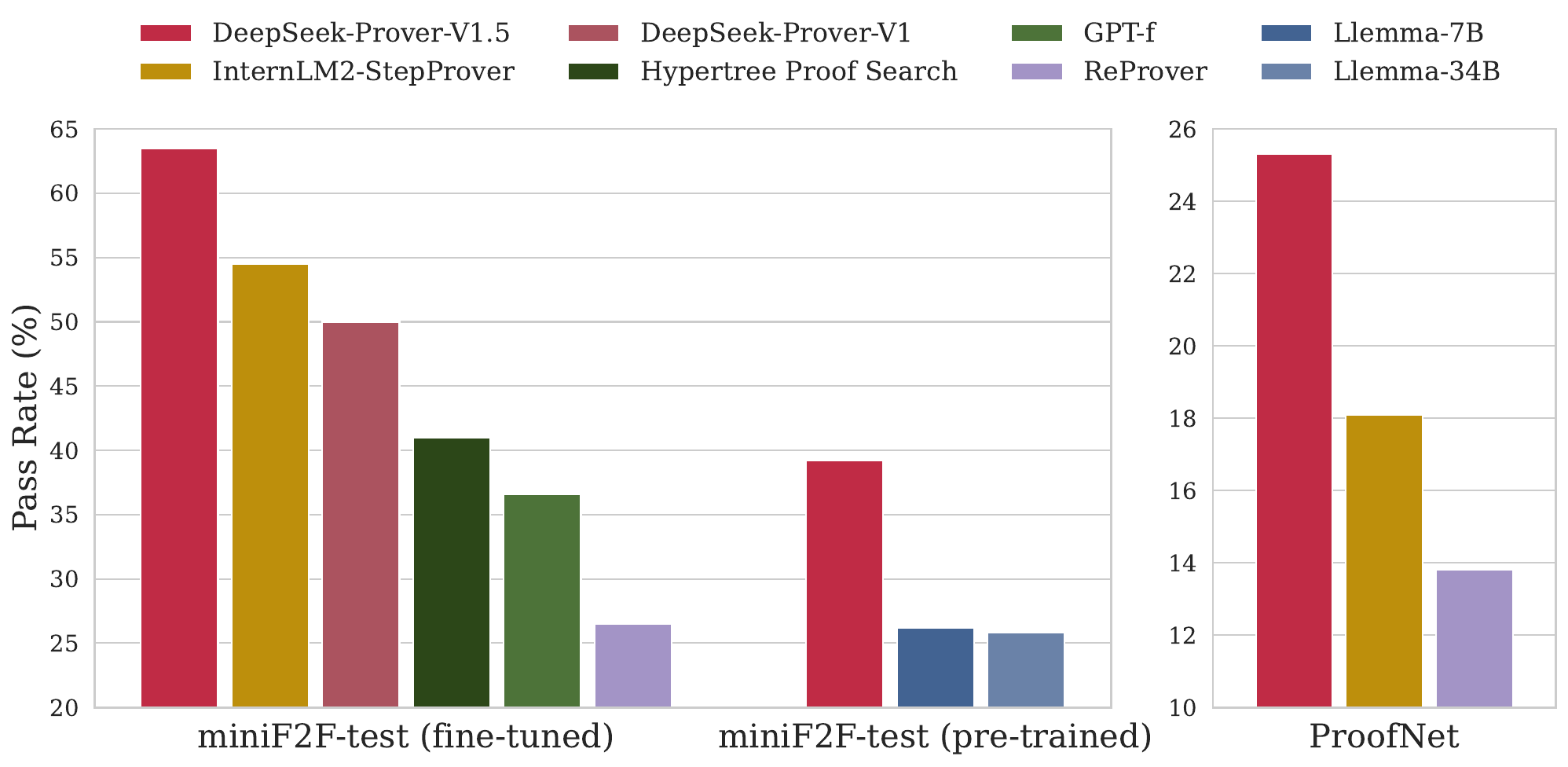}
  \caption{Pass rates of models on formal theorem proving benchmarks in Lean 4: the high school level miniF2F-test benchmark~\citep{zheng2021minif2f} and the undergraduate level ProofNet benchmark~\citep{azerbayev2023proofnet}. We compare both the pre-trained and fine-tuned versions of \thiswork~with strong baselines.
  }
  \label{fig:performance}
\end{figure}


\newpage

\section{Introduction}

Recent advancements in large language models have significantly influenced mathematical reasoning and theorem proving in artificial intelligence. Despite notable progress in natural language domains, language models still encounter substantial challenges in formal theorem proving, \eg~using Lean \citep{moura2021lean} and Isabelle \citep{paulson_isabelle_1994}, which requires rigorous derivations satisfying formal specifications of the verification system.
Even advanced models like GPT-4 \citep{OpenAI} struggle with complex formal proofs, underscoring the intricate nature of both the coding and the mathematics involved.
A formal theorem proving model must not only grasp the syntax and semantics of formal systems like the Lean theorem prover but also align abstract mathematical reasoning with precise formal representation.

Language models in formal theorem proving typically employ two strategies: proof-step generation \citep{polu2020generative, jiang2022thor, lample2022hypertree, yang2024leandojo, wu2024lean} and whole-proof generation \citep{jiang2022draft, zhao2023decomposing, wang2023lego}. Proof-step generation predicts each subsequent tactic and verifies it using the formal verifier to obtain updated information about the current tactic state, often utilizing tree search techniques to construct valid proofs. In contrast, whole-proof generation is computationally efficient, which produces an entire proof code based on the theorem statement, requiring less communication budget to coordinate between the prover model and the formal theorem verifier. While \lastwork~\citep{xin2024deepseek} has achieved state-of-the-art results in Lean 4 with whole-proof generation, this paradigm presents its unique challenges. It requires long-horizon sequence prediction without access to intermediate tactic states, and future tactics depend on these hidden results. In Lean's tactic mode, proofs are constructed through a sequence of tactics that transform the proof state. This sequential nature introduces the risk of compounding errors \citep{ross2011reduction}, where a single misinterpretation can lead to significant deviations from a valid proof path. More specifically, the auto-regressive model may have incorrect believes on intermediate tactic states when generating long proofs.

To seamlessly integrate intermediate tactic states in proof-step generation while maintaining the simplicity and computational efficiency of whole-proof generation, we have developed a unified approach in \thiswork. This method combines the strengths of both proof-step and whole-proof generation techniques through a truncate-and-resume mechanism. The process begins with standard whole-proof generation, where the language model completes the proof code following the theorem statement prefix. The Lean prover then verifies this code. If the proof is correct and complete, the procedure terminates. If an error is detected, the code is truncated at the first error message, and any subsequent code is discarded. The successfully generated proof code is then used as a prompt for the generation of next proof segment. To enhance the accuracy of the model's new completions, we append the latest state from the Lean 4 prover as a comment at the end of the prompt. Notably, our method is not restricted to resuming from the last successfully applied tactic. We integrate the truncate-and-resume mechanism into Monte-Carlo tree search \citep[MCTS;][]{coulom2006efficient} in which the truncation points are scheduled by the tree search policy.
In addition, we propose a novel reward-free exploration algorithm for MCTS to address the reward sparsity issue of proof search.
We assign the tree search agent intrinsic motivation, \aka~curiosity \citep{schmidhuber2010formal}, to extensively explore the tactic state space.
These algorithmic modules extend the functionality of our whole-proof generation model to become a flexible tool for interactive theorem proving, which can effectively utilize the proof assistant feedback and generate diverse solution candidates.

\begin{figure}[htb]
   \centering
   \includegraphics[width=0.75\textwidth]{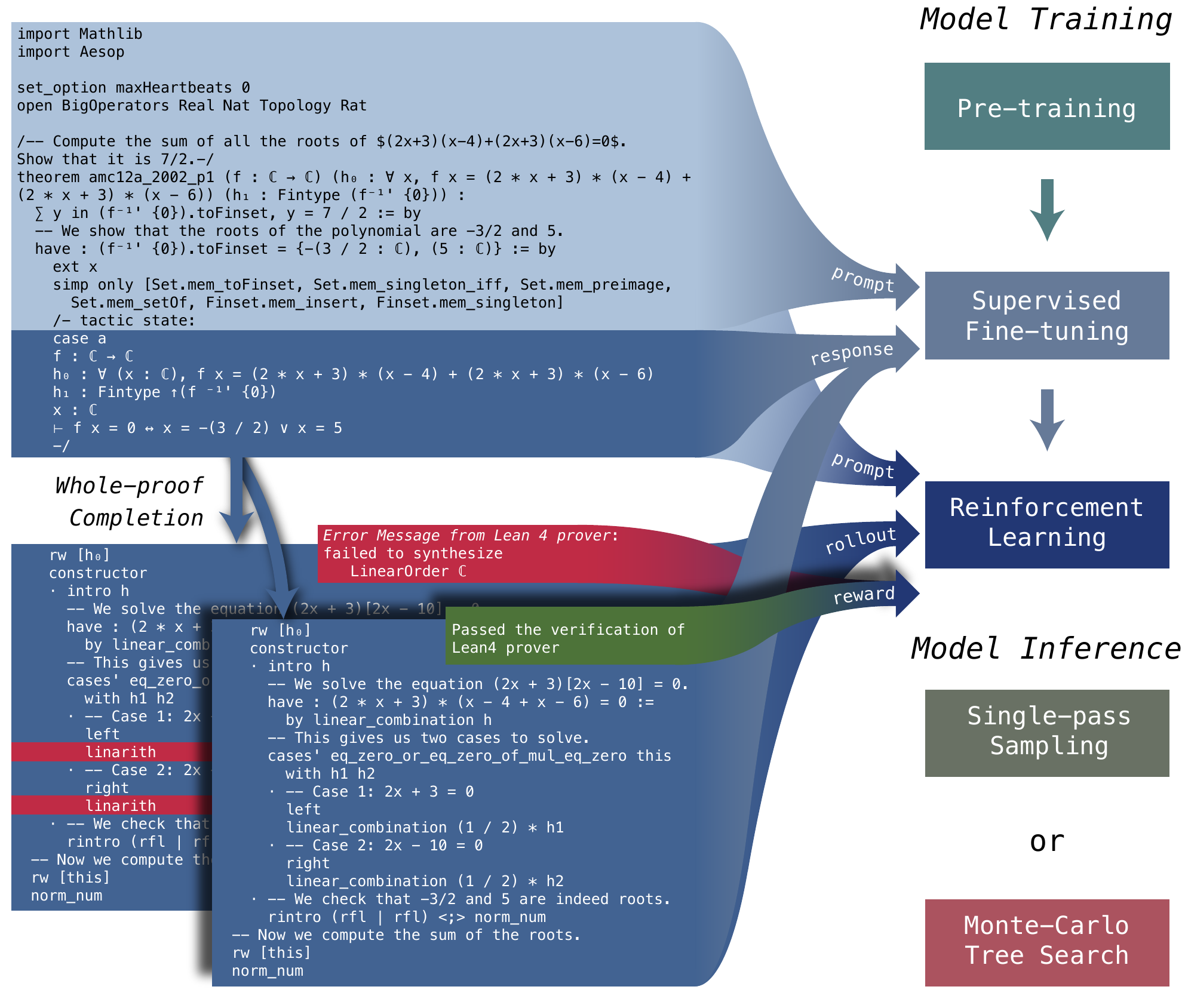}
   \caption{\textbf{Overall Framework.} \thiswork~is trained through pre-training, supervised fine-tuning, and reinforcement learning. During supervised fine-tuning, the pre-trained model receives an incomplete theorem proof ending with a tactic state comment keyword. The model is trained to predict the content of this tactic state (auxiliary objective) and complete the subsequent proof steps (main objective). In the reinforcement learning stage, given an incomplete theorem proof and ground-truth tactic state from the Lean prover, we roll out the fine-tuned model to generate multiple proof candidates, which are then verified by the Lean prover. The verification results for these candidates are used as binary (0-1) rewards to further optimize the model and enhance its alignment with the formal specifications of the verification system. For model inference, we offer two alternatives: single-pass sampling and Monte-Carlo tree search.}
  \label{fig:overview_training}
\end{figure}

\subsection{Contributions}

We present a comprehensive framework for developing a language model-based formal mathematics prover, integrating several key components: large-scale mathematical pre-training, formal mathematics corpus construction and augmentation, online reinforcement learning from proof assistant feedback, and a tree search methodology for long-term planning in theorem proving. The pre-trained model, supervised fine-tuned model, and reinforcement learning model, along with the code for the Monte-Carlo tree search algorithm, are publicly available for further research and application.

\begin{itemize}
\item \textbf{Pre-Training}: We enhance our base model's capabilities in formal theorem proving and mathematical reasoning by further pre-training on high-quality mathematics and code data, with a focus on formal languages such as Lean, Isabelle, and Metamath.
\item \textbf{Supervised Fine-Tuning}: We improve the Lean 4 code completion dataset by implementing two data augmentation techniques. First, we use DeepSeek-Coder V2 236B \citep{zhu2024deepseek} to annotate natural language chain-of-thought comments alongside Lean 4 code, aligning formal theorem proving with natural language reasoning. Second, we insert intermediate tactic state information within the Lean 4 proof code, enabling our model to leverage compiler feedback effectively. The resulting dataset is then used to fine-tune the pre-trained model.
\item \textbf{Reinforcement Learning}: We employ the GRPO algorithm \citep{shao2024deepseekmath} to perform reinforcement learning from proof assistant feedback (RLPAF) on the supervised fine-tuned model. Verification results from the Lean prover serve as reward supervision, enhancing the model's alignment with the formal specifications of the verification system.
\item \textbf{Monte-Carlo Tree Search}: We advance the tree search method in formal theorem proving by introducing a novel abstraction and a corresponding search algorithm. Our truncate-and-resume mechanism acts as a state-action abstraction, seamlessly integrating the tree search process into the whole-proof generation framework. We present \treesearch, an innovative Monte-Carlo tree search algorithm that leverages the RMax \citep{brafman2002r} strategy to tackle exploration challenges in sparse-reward proof search problems. By assigning intrinsic rewards, this algorithm encourages the prover agent to generate diverse planning paths, thereby fostering extensive exploration of the proof space.
\end{itemize}

\subsection{Summary of Evaluations and Metrics}

\begin{itemize}
\item \textbf{miniF2F}: In the single-pass whole-proof generation setting, \thiswork~achieved a pass rate of \textbf{60.2\%} on the test set of miniF2F, marking a significant improvement of absolute 10.2 percentage points over \lastwork's 50.0\%. Incorporating tree search techniques further elevated the pass rate to a new state-of-the-art \textbf{63.5\%}.
\item \textbf{ProofNet}: \thiswork~also demonstrated strong performance in the single-pass whole-proof generation setting for ProofNet, with pass rates of \textbf{21.6\%} on the validation set and \textbf{23.7\%} on the test set. The integration of tree search techniques further enhanced these results, achieving new state-of-the-art pass rates of \textbf{25.4\%} on the validation set and \textbf{25.3\%} on the test set.
\end{itemize}

\section{Model Training}

\subsection{Pre-training}

To enhance our language model's proficiency in generating formal proofs and reasoning through mathematical language, we further pre-train our base model~\citep{shao2024deepseekmath}. This refinement involved training on high-quality datasets that include both code and natural language mathematical content. We specifically focused on formal languages widely used in proof assistants, such as Lean, Isabelle, and Metamath. We designate this improved model as \thiswork-Base.

\subsection{Supervised Fine-tuning}

In this section, we explore the methodology and processes involved in the supervised fine-tuning (SFT) of \thiswork. Specifically, we augment the proof dataset from \lastwork~by adding detailed explanatory comments. This enhancement aims to improve the alignment between natural language descriptions and Lean 4 code, thereby facilitating better formal mathematical reasoning. Additionally, we incorporate intermediate tactic state information as an auxiliary prediction task to support the truncate-and-resume mechanism used in the Monte-Carlo Tree Search process. We refer to the resulting model as \thiswork-SFT.

\vspace{-0.1in}

\paragraph{Data Curation.}
We develop a comprehensive Lean 4 code completion dataset for the supervised fine-tuning. This dataset includes synthetic proof code derived from a wide range of formal theorems. These theorems are sourced from various projects, such as the standard Lean 4 math library Mathlib4 \citep{mathlib}, synthetic theorems from \lastwork~\citep{xin2024deepseek} and Lean Workbook \citep{ying2024lean}, and validation sets from the miniF2F \citep{zheng2021minif2f} and ProofNet \citep{azerbayev2023proofnet} benchmarks. To augment the formal proof data, we employed an expert iteration process \citep{polu2020generative}. This involves generating proofs using the language model, verifying the generated proof data, retraining the model with the verified data, and then using the optimized model to generate additional proof data. Between each iteration, we use DeepSeek-Coder V2 236B \citep{zhu2024deepseek} to annotate the thought process before the proof code as comments. Finally, we tailor these data for the truncate-and-resume mechanism for Monte-Carlo Tree Search (details in Section~\ref{sec:tree-abstraction}). The resulting proof dataset consists of 9,645k sequences.

\vspace{-0.1in}

\paragraph{Thought-augmented Proof Generation. } \label{sec:thought-augmented_proof_generation}
In \lastwork, we identified a significant gap between problem-solving strategies in natural language and theorem proving in Lean. In natural language, models generate detailed deduction steps to construct proofs, whereas in Lean, they often rely on a sequence of high-level tactic calls to brute-force solutions. These high-level tactics, while effective, obscure their internal workings and outcomes, hindering the model's ability to resolve complex proof goals with structured mathematical reasoning. To address this issue, we develop an approach that incorporates natural language reasoning before generating theorem proof code. Similar to Lean-STaR \citep{lin2024lean}, which performs isolated chain-of-thought reasoning \citep{wei2022chain, feng2024towards} before each proof step, our method integrates this reasoning directly as comments within the proof code. We use the DeepSeek-Coder V2 236B \citep{zhu2024deepseek} to enhance existing data in \lastwork~in two ways: first, by inserting a complete natural language solution at the beginning of the proof block, and second, by alternately inserting specific natural language steps for corresponding Lean tactics. Training the model with this data format enforces it to propose complete mathematical reasoning at the beginning of the proof block and detailed step planning before each tactic. This approach successfully develops new behaviors, employing delicate mathematical thinking to guide the generation of tactics. In the training data, two distinct guiding prompts are used to differentiate between the CoT (Chain of Thought) mode and the non-CoT mode for proof code completion. Examples of input and output in both modes can be found in Appendix~\ref{app:promt_examples}.

\vspace{-0.1in}

\paragraph{Prompt Augmentation with Tactic State Information. } \label{sec:sft-tactic-state}
To implement the truncate-and-resume mechanism for Monte-Carlo Tree Search, we needed to extract tactic information from the code generated by the model. We enhanced the Lean REPL \citep[Read-Eval-Print Loop;][]{repl} with data extraction tools from the LeanDojo \citep{yang2024leandojo} project. This allowed us to extract tactic information in triples, which include the position of each tactic, as well as the tactic states before and after its application. This information helps us identify the specific tactic code that triggers verification errors (used in the expansion step for tree search, see Section~\ref{para:expansion}). For each tactic in a generated valid formal proof, we insert the tactic state returned by the verifier as a comment "/- tactic state: ... -/". During training, we use all tokens following "/- tactic state: " as responses to calculate the supervised fine-tuning loss, while the tokens before this comment is used as prompts and do not contribute to the training loss calculation.

\vspace{-0.1in}

\paragraph{Training Setting. }
We conduct supervised fine-tuning based on the pre-trained model and train for 9B tokens, using a batch size of 2,048 and a constant learning rate of 1e-4. The training process begins with 100 warm-up steps to stabilize the learning dynamics. Training examples are randomly concatenated to form sequences, with a maximum context length of 4,096 tokens.

\subsection{Reinforcement Learning from Proof Assistant Feedback}

Reinforcement learning (RL) has been proven effective in enhancing the mathematical reasoning capabilities of supervised fine-tuned language models~\citep{shao2024deepseekmath}. To further advance \thiswork-SFT, we incorporate a reinforcement learning phase, resulting in the model \thiswork-RL. This phase leverages RL to enhance performance based on verification feedback from the Lean 4 prover. The specifics of this RL process are detailed below.

\paragraph{Prompts.}
In the reinforcement learning stage, we use a subset of theorem statements from the supervised fine-tuning dataset as training prompts. We select theorems for which \thiswork-SFT has a moderate success rate in generating correct proofs upon multiple attempts. This ensures that the model has room for improvement while still being able to receive positive feedback. After filtering, we retain approximately 4.5k unique theorem statements. Each theorem is prefixed with both CoT and non-CoT guiding prompts to enhance the model's proof generation capabilities in both modes.

\paragraph{Rewards.}
When training LLMs via RL, a trained reward model typically provides feedback signals. In contrast, formal theorem proving benefits from the rigorous verification of generated proofs by proof assistants, offering a significant advantage. Specifically, each generated proof receives a reward of 1 if verified as correct, and 0 otherwise. While this binary reward signal is accurate, it is also sparse, especially for theorems that are challenging for the supervised fine-tuned model. To mitigate this sparsity, we select training prompts that are challenging yet achievable for the supervised fine-tuned model, as described above.

\paragraph{Reinforcement Learning Algorithm.}
We employ the Group Relative Policy Optimization \citep[GRPO;][]{shao2024deepseekmath} as our RL algorithm, which has demonstrated superior effectiveness and efficiency compared to PPO \citep{schulman2017proximal}, primarily because it eliminates the necessity of training an additional critic model. Specifically, GRPO samples a group of candidate proofs for each theorem prompt and optimizes the model based on the relative rewards of the outputs within the group. Our prompt selection strategy is designed to likely include both correct and incorrect proofs among the candidates, aligning well with the group-relative nature of GRPO and thereby enhancing the training process.

\paragraph{Training Setting. }
We conduct RL training based on the SFT model, which serves as both the initial model and the reference model for imposing the Kullback-Leibler (KL) divergence penalty. We use a constant learning rate of 5e-6, and the KL penalty coefficient is set to 0.02. For each theorem, we sample a group of 32 candidate proofs, with maximum length set to 2,048. The training batch size is configured to 512.

\subsection{Evaluation} \label{sec:sft-eval}

\paragraph{Benchmarks.}
We evaluate theorem-proving performance on the following benchmarks to compare model capabilities after each training stage:
\begin{itemize}
    \item \textbf{MiniF2F}~\citep{zheng2021minif2f} focuses on formal problem-solving skills for high-school level exercises and competitions, such as AMC, AIME, and IMO, with an emphasis on algebra and number theory. The benchmark includes 244 validation and 244 test problems, originally in Lean 3 and manually converted to Lean 4.9.0, based on the version provided by \citet{minif2f_solution}.
    \item \textbf{ProofNet}~\citep{azerbayev2023proofnet} evaluates formal theorem-proving capabilities at the undergraduate level in mathematics. It comprises 185 validation and 186 test problems from widely-used undergraduate textbooks, covering real and complex analysis, linear algebra, abstract algebra, and topology. These problems were initially in Lean 3 and manually converted to Lean 4.9.0.
\end{itemize}

\begin{figure}[t]
  \begin{minipage}{0.55\textwidth}
    \centering
    \includegraphics[width=0.98\textwidth]{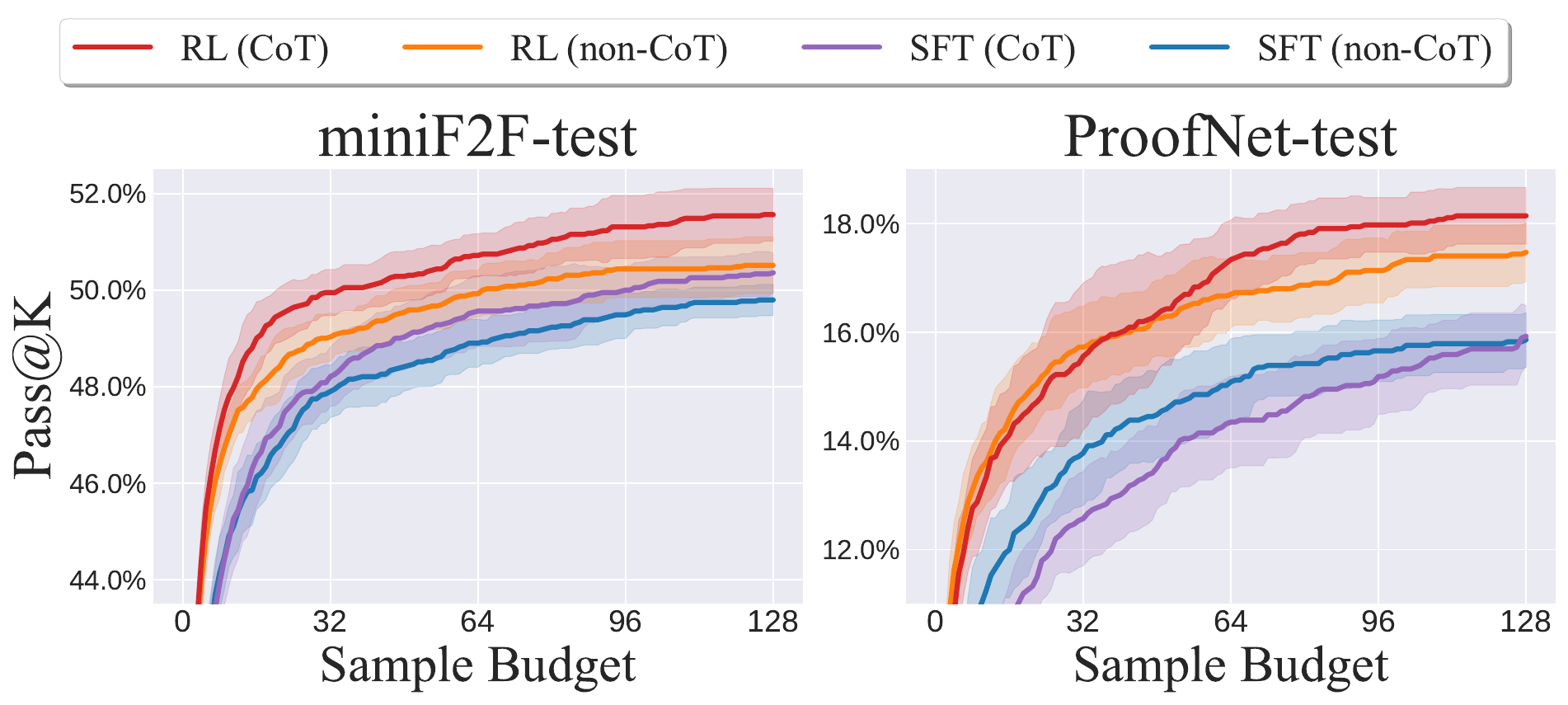}
  \end{minipage}
  \begin{minipage}{0.45\textwidth}
    \centering
    \footnotesize
    \begin{tabular}{lcc}
    \toprule 
        \multirow{2}{*}{Model} & \multicolumn{2}{c}{Pass@128} \\
        & miniF2F-test & ProofNet-test \\
        \toprule
        Base (3-shot) & $29.7\% \pm 0.5\%$ & $9.7\% \pm 0.7\%$ \\
        \midrule
        SFT (non-CoT) & $49.8\% \pm 0.3\%$ & $15.9\% \pm 0.5\%$ \\
        SFT (CoT) & $50.4\% \pm 0.4\%$ & $15.9\% \pm 0.6\%$ \\
        \midrule
        RL (non-CoT) & $50.5\% \pm 0.6\%$ & $17.5\% \pm 0.5\%$\\
        RL (CoT) & $51.6\% \pm 0.5\%$ & $18.2\% \pm 0.5\%$ \\
        \bottomrule
    \end{tabular}
  \end{minipage}
  \caption{\textbf{Comparison of model capabilities at different training stages.} "CoT" and "non-CoT" refer to evaluations using two guiding prompts. The shaded region represents the range of standard deviations around the mean values. The notation $\mu\pm\sigma$ indicates the average accuracy $\mu$ and the standard deviation $\sigma$.}
  \label{fig:sample128}
\end{figure}

\paragraph{Prompting Configurations.}
For each proof attempt of \thiswork-Base, we independently sample three proof demonstrations from the validation set to construct the few-shot prompts. For the miniF2F benchmark, we use human-written proofs from \citet{minif2f_solution}, while for the ProofNet benchmark, we use correct proofs generated by \thiswork-RL as few-shot demonstrations. For \thiswork-SFT and \thiswork-RL, we employ two types of guiding prompts: one that encourages chain-of-thought (CoT) reasoning before each proof step, and one that does not (non-CoT). Detailed examples are provided in Appendix~\ref{app:promt_examples}.

\paragraph{Metric.}
We evaluate theorem-proving performance using the pass@$K$ accuracy metric, which measures the model's success in generating a correct proof within $K$ attempts. Each model is deployed on a single A100-40G GPU, utilizing the vLLM framework~\citep{kwon2023efficient} for sample generation. The sampling parameters are set with a temperature of 1, a top-p value of 0.95, and a maximum token limit of 2,048. The generated proofs are then verified using the Lean 4 theorem prover. For this verification, we import Mathlib4 \citep{mathlib} and Aesop \citep{limperg2023aesop} to access predefined premises and tactics. The verification process is subject to a time limit of 300 seconds.

\paragraph{Comparison across Training Stages.}
Figure~\ref{fig:sample128} presents a comparative analysis of each training stage on the miniF2F and ProofNet datasets. Our base model, \thiswork-Base, achieves a notable pass rate, solving nearly one-third of the problems on the test set of the miniF2F benchmark using 3-shot prompting. The supervised fine-tuning stage, resulting in \thiswork-SFT, significantly outperforms the base model, with Pass@128 accuracy increasing by approximately two-thirds on miniF2F and doubling on ProofNet. The subsequent reinforcement learning stage further enhances the model's performance, improving Pass@$K$ accuracy across all values of $K$. In contrast to findings in natural language mathematics, such as those reported in DeepSeekMath~\citep{shao2024deepseekmath}, where reinforcement learning primarily boosts the correct response from TopK, we observe a genuine enhancement of fundamental capabilities in formal theorem proving. This improvement is evident not only with a small sample budget but also remains stable as the sample budget increases. This conclusion is further supported by later Monte-Carlo Tree Search experiments with larger sample budgets, as discussed in Section~\ref{sec:large-train-eval}.

\paragraph{Comparison between CoT and non-CoT.}
We compare the performance of non-CoT and CoT generation modes for both \thiswork-SFT and \thiswork-RL. The results in Figure~\ref{fig:sample128} demonstrate that the CoT mode consistently outperforms the non-CoT mode across most settings. Specifically, \thiswork-RL, leveraging these enhanced theorem-proving patterns, achieves superior performance on both benchmarks, with an average accuracy of $51.6\%$ on miniF2F and $18.2\%$ on ProofNet. The integration of natural language reasoning in CoT mode significantly enhances the planning and execution of formal proof writing. For a detailed comparison of proof strategies with and without the use of natural language chain-of-thought, refer to the examples provided in Appendix~\ref{app:promt_examples}.

\section{Exploration-oriented Monte-Carlo Tree Search} \label{sec:tree-search}

\subsection{Tactic-level Tree Abstraction} \label{sec:tree-abstraction}

\begin{figure}[t!]
   \centering
   \includegraphics[width=0.8\textwidth]{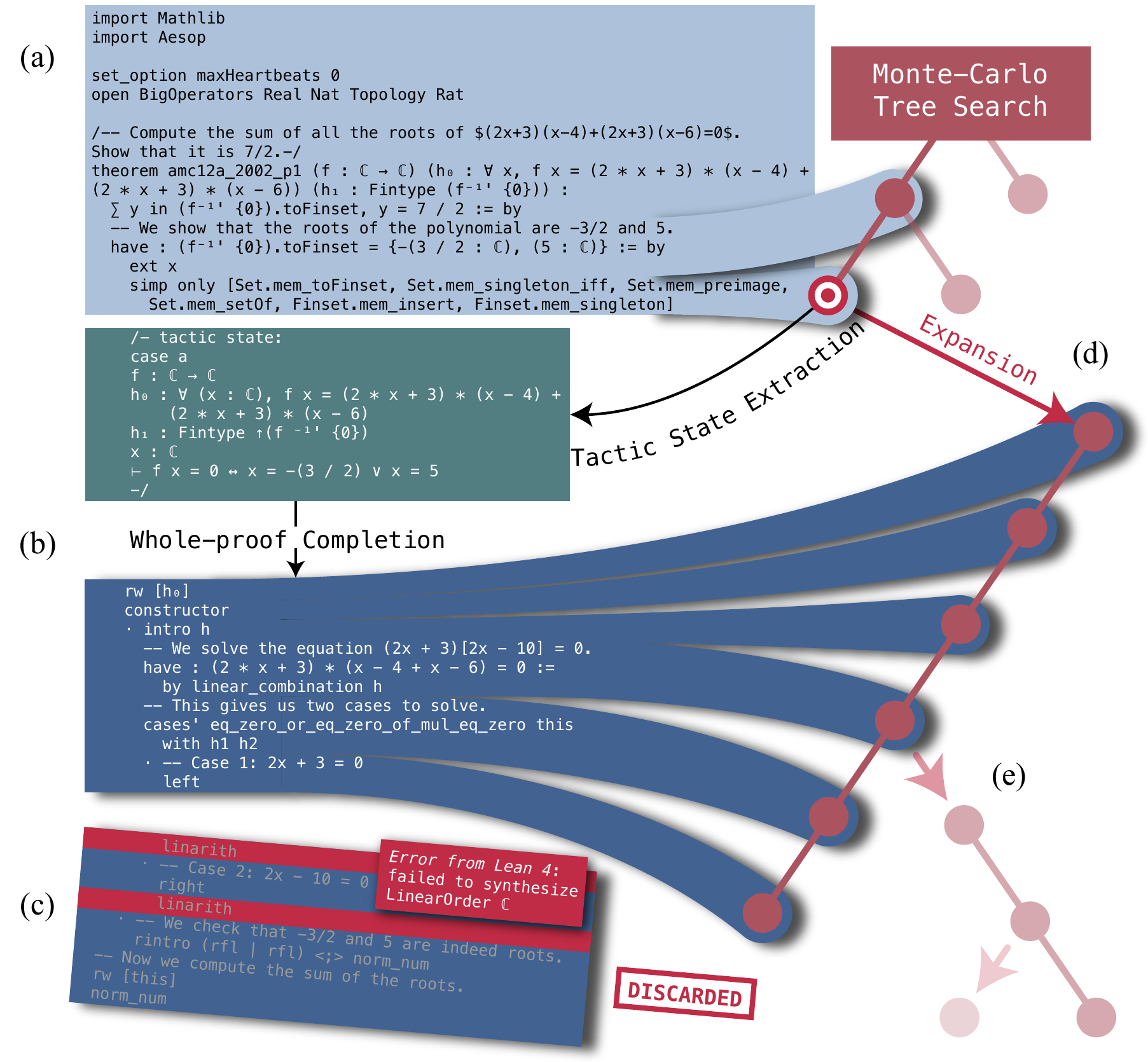}
  \caption{\textbf{Truncate-and-Resume Mechanism in the Expansion Step of MCTS.} (a) After selecting a node, we trace its corresponding incomplete proof code prefix, which includes the file header, initial statement, and successfully applied tactics from the ancestor nodes. (b) The language model then generates the subsequent proof based on this prefix along with a comment block containing the current tactic state. (c) The combined proof code (prefix and newly generated code) is verified by the Lean 4 prover. If no errors are found, the tree-search procedure terminates. If errors are detected, we truncate the newly generated code at the first error message, discard the subsequent code, and parse the successful portion into tactics. (d) Each tactic is added as a new node in the search tree, extending a chain of descendants beneath the selected node. (e) Once the tree updates are complete, the next iteration of expansion begins by selecting an alternative candidate node, which is not limited to leaf nodes. This process repeats until a correct proof is found or the sample budget is exhausted.}
  \label{fig:overview_mcts}
\end{figure}

To implement the tree search method in the whole-proof generation setting, we introduce a proof tree abstraction to define the tailored state and action space, leveraging a truncate-and-resume mechanism.
Roughly following the paradigm of \citet{yao2023tree}, we begin by decomposing an incomplete proof into a sequence of tree nodes that correspond to individual proof steps, and then we utilize the partial content stored in these tree nodes to continue the proof generation process.
Figure~\ref{fig:overview_mcts} illustrates the process of constructing a proof search tree from whole-proof generation.

\paragraph{Truncate: Proof Decomposition into Tree Nodes.}
We construct the proof search tree at the tactic level, where each tree edge represents a single transition step of the tactic state. Initially, we submit the entire proof the model generated to the Lean prover to parse it into tactics. We then truncate the proof at the earliest verification error, ensuring that all subsequent tactic codes can be successfully applied to advance the proof towards the desired theorem. The tactic codes are segmented into several code fractions, each containing a valid tactic code and its associated chain-of-thought comments, corresponding to a single tree edge that represents a tactic state transition. Through this abstraction, each tactic code is converted into a series of tree nodes, forming a path from the root to a specific node.

\paragraph{Resume: Proof Generation from a Tree Node.}
In Lean 4, different tactics can lead to the same tactic state, meaning each node in our proof tree can correspond to various tactic codes that achieve the same outcome. To handle this, we store a set of these equivalent tactic codes at each node. When the tree search agent expands a node, it randomly selects one tactic to use as a prompt for the language model. This prompt includes the incomplete proof code ending with the chosen tactic and the tactic state information from the Lean prover as a comment block. The fine-tuned model (see Section \ref{sec:sft-tactic-state}) has been trained to recognize and utilize this format, using the incomplete code augmented with tactic state comments to guide subsequent proof generation.

\subsection{Interactive Theorem Proving via Monte-Carlo Tree Search}
\label{sec:mcts}
Our proof search tree is developed using the standard Monte-Carlo Tree Search (MCTS) paradigm \citep[MCTS;][]{coulom2006efficient, browne2012survey}, which iteratively applies four steps: \textit{Selection}, \textit{Expansion}, \textit{Simulation}, and \textit{Backpropagation}.
We integrate the \textit{Simulation} step into \textit{Expansion} because our whole-proof generation model inherently performs a rollout from the expanded node.
The detailed design of the algorithm workflow is as follows.

\paragraph{Selection.}
The selection step, \aka the tree policy, starts from the root node and traverses downward to identify a promising node for expansion.
The objective of this algorithmic step is to trade off between exploration and exploitation \citep{kocsis2006bandit}.
The tree policy at a tree node $s$ is computed by selecting the action that maximizes the value from the set of valid operations:
\begin{align} \label{eq:tree_policy}
    TreePolicy(s) = \mathop{\arg\max}_{a\in Children(s)\cup\{\oslash\}} Q_{UCB}(s, a),
\end{align}
where the action $a$ can be either moving to a child node, denoted by $a\in Children(s)$, or expanding the current node $s$, denoted by a special token $a=\oslash$.
This approach uses a technique called \textit{virtual node} \citep{wang2023dt}, which assigns each node an imaginary child to represent the selection of the current node $s$ for expansion. 
It enables the tree search agent to continually expand non-leaf nodes, as the action space is supported by a generative model whose output scope cannot be determined by a fixed number of trails.
The value estimation $Q_{UCB}(s,a)$ of performing action $a$ on node $s$ is composed by two components:
\begin{align}
    \forall a\in Children(s)\cup\{\oslash\},\quad Q_{UCB}(s, a) &= \underbrace{Q(s, a)}_{\text{Exploitation}} + ~~~ \underbrace{UCB(s, a)}_{\text{Exploration}},
\end{align}
where $Q(s, a)$ denotes a sample-based estimation of action values derived from the selection history, functioning as the exploitation component that retrieves high-value candidates from previous trials. $UCB(s, a)$ denotes the exploration bonus computed by upper confidence bounds \citep[UCB;][]{auer2002using}, which diminishes with the repeated execution of the state-action pair $(s,a)$. More specifically, $Q_{UCB}(s, a)$ stands for an optimistic estimation of $Q(s, a)$ and can serve as an upper bound with high probability. We defer the discussion of detailed settings of node values and UCB bonus to Section~\ref{sec:rmax}.

\paragraph{Expansion.}\label{para:expansion}
The next step is invoking the proof generation model to expand the node nominated by the selection phase.
Resuming the incomplete proof codes stored on the node designated for expansion, we perform whole-proof generation to propose a series of subsequent tactics and submit the generated proof to Lean prover for verification.
Such a trial of proof completion is equivalent to conducting a single rollout of simulation within the standard MCTS framework.
When the verification result indicates the proof is complete, the search procedure is ready to be terminated, having found a new proof of the desired theorem.
Otherwise, we parse the verification feedback and truncate the generated proof to the assertion of the earliest verification error.
The remaining tactics are transformed into a path of nodes to be merged into the search tree (see Figure~\ref{fig:overview_mcts}).
It is important to note that, because we use the whole-proof generation setting—where the output is an entire proof consisting of a sequence of tactics, rather than just the next tactic—our expansion procedure may insert a path of tree nodes into the search tree during each iteration. This differs from the conventional MCTS designed for competitive games, which typically expands only one layer of children nodes per iteration \citep{silver2016mastering, silver2018general, schrittwieser2020mastering}.

\paragraph{Backpropagation.}
The final phase of each tree search iteration is to update value statistics along the selection trajectory from the root to the expanded node, \ie, updating the values associated with the tree policy stated in Eq.~\eqref{eq:tree_policy}.
Let $\tau=\{(root, s^{(1)}), (s^{(1)}, s^{(2)}), (s^{(2)}, s^{(3)}), \dots, (s^{(|\tau|-1)}=s_t, \oslash)\}$ denote the selection trajectory of $t$-th iteration that ends with $s_t$ as the expanding node.
We update $Q_{UCB}(s,a)$ for all $(s,a)\in\tau$ by taking the most recent trajectory reward $R(\tau)$ into account (details refer to Eq.~\eqref{eq:ducb}).
The extrinsic source of rewards comes from the compiler feedback, specifically assigning a reward of $R_{\text{extrinsic}}(\tau)=1$ for completed proofs and $R_{\text{extrinsic}}(\tau)=0$ for unsolved ones.
In Section~\ref{sec:rmax}, we will introduce an intrinsic reward mechanism to augment the reward assignment that enhances the agent's incentive for exploration.

\subsection{Intrinsic Rewards for Monte-Carlo Tree Search} \label{sec:rmax}

In the search problem of formal theorem proving, the extrinsic rewards are extremely sparse, \ie, the search agent only obtains non-zero rewards when the proof is completely solved.
More specifically, the proof search process forms a tree structure with only a narrow set of leaves delivering non-zero rewards, which matches a famous hard-exploration case \citep{krishnamurthy2016pac} in the literature of statistical reinforcement learning.
To promote exploration in sparse-reward sequential decision making, one classical paradigm is constructing intrinsic rewards \citep{schmidhuber2010formal} that encourage the agent to not only optimize extrinsic rewards but also acquire general information about the interactive environment \citep{bellemare2016unifying, houthooft2016vime, pathak2017curiosity, burda2019exploration}.
In this section, we present our intrinsic-reward-driven exploration algorithm, \textit{RMax applied to Tree Search} (\treesearch), to incorporate reward-free exploration in the proof search problem.

\paragraph{RMax applied to MCTS.}
We adopt RMax \citep{brafman2002r}, a classical exploration mechanism, to construct intrinsic rewards for Monte-Carlo tree search.
The core idea of RMax is to explore a broad coverage of the state space.
The agent awards itself a maximal amount of reward upon reaching an unseen state.
In the context of proof search, where no extrinsic rewards are provided until the proof is completed, our algorithmic procedure resembles ZeroRMax \citep{jin2020reward}, in which the agent's exploration is driven solely by intrinsic rewards, \ie, setting $R(\tau)=R_{\text{intrinsic}}(\tau)$.
The intrinsic reward of a tree expansion step is determined by whether a new node is added to the search tree,
\begin{align}\label{eq:rmax_reward}
    R_{\text{intrinsic}}(\tau) = \mathbb{I}\left[\text{at least one new node is added to the search tree}\right],
\end{align}
where $\tau$ denotes the most recent selection trajectory that requires a reward assignment for backpropagation.
This exploration strategy prioritizes the expansion of nodes where the prover model generates tactics that lead to a diverse range of tactic states.
As multiple Lean codes can result in the same transition of intermediate states, this heuristics can potentially reduce redundant generation and improve sample efficiency.

\paragraph{UCB for Non-stationary Rewards.}
The common setting of UCB exploration bonus for Monte-Carlo tree search is using UCB1 \citep{auer2002finite}:
\begin{align} \label{eq:ucb1}
    Q_{UCB1}(s, a) &= \frac{W(s,a)}{N(s, a)} + \sqrt{\frac{2\ln\sum_{a'}N(s,a')}{N(s,a)}}, \\
    W(s,a) &= \textstyle{\sum_{\tau\in\Gamma(s,a)}}~R(\tau), \\
    N(s,a) &= \left|\Gamma(s,a)\right|,
\end{align}
where $\Gamma(s,a) = \{\tau\mid (s,a)\in\tau\}$ denotes the list of tree-policy trajectory $\tau$ containing $(s,a)$ as an intermediate selection step.
To facilitate discussions, we organize the list $\Gamma(s,a)=\{\tau_1,\tau_2,\cdots\}$ such that newly collected trajectories have larger subscript indices.
In this work, we propose to use an alternative variant of UCB method.
Note that the derived intrinsic reward in Eq.~\eqref{eq:rmax_reward} is a non-stationary reward signal whose expected value decays with the progress of exploration.
That is because it becomes definitely harder to discover new nodes with unseen tactic states as the search tree expands through sophisticated exploration.
To tackle the non-stationarity, we consider \textit{discounted upper confidence bounds} \citep[DUCB;][]{garivier2011upper}, which uses a discount factor $\gamma\in(0,1)$ to smoothly drop those outdated feedback records:
\begin{align} \label{eq:ducb}
    Q_{DUCB}(s, a) &= \frac{W_\gamma(s,a)}{N_\gamma(s, a)} + \sqrt{\frac{2\ln\sum_{a'}N_\gamma(s,a')}{N_\gamma(s,a)}}, \\
    W_\gamma(s,a) &= \textstyle{\sum_{t=1}^{N(s,a)}}~\gamma^{N(s,a)-t}R(\tau_t), \\
    N_\gamma(s,a) &= \textstyle{\sum_{t=0}^{N(s,a)-1}}~\gamma^t,
\end{align}
where newly received feedback would be assigned a larger weight in the value estimation.
In practice, we set $\gamma=0.99$.
Note that the role of discount factor $\gamma$ in DUCB differs from its role in value iteration for infinite-horizon MDPs.
The discounting is applied to tree search iterations rather than to the action-step horizon within a single trajectory.

\subsection{Parallelization of Monte-Carlo Tree Search}

To enhance the efficiency of Monte-Carlo Tree Search (MCTS), we implement several established parallelization techniques as described by \citet{chaslot2008parallel}.
\begin{itemize}
    \item \textbf{Root Parallelization:} We deploy 256 MCTS runners per node, with one language model per GPU and a batch size of 512 for proof generation. The Lean prover is invoked through REPL and executed on a cluster with thousands of CPU cores, where each proof verification task is handled by an individual process, created and terminated in a sandbox. Both proof generation by language models and verification by Lean provers are handled asynchronously. This setup allows MCTS runners to perform concurrent tree search operations, significantly accelerating the process.
    \item \textbf{Tree Parallelization:} We manage each search tree with 32 thread workers to parallelize the tree iteration steps. This method effectively schedules and balances the tasks of proof generation and Lean verification. Each thread worker iteratively performs the tree search loop by selecting a candidate node for expansion, invoking the language model to generate the proof, verifying the generated proof with the Lean prover, and performing backpropagation.
    \item \textbf{Virtual Loss:} To encourage diverse node selection among concurrent thread workers, we assign a virtual reward $R(\tau)=0$ for ongoing iterations. This involves backpropagating a reward of $0$ temporarily and updating it to the true reward upon completion. This strategy promotes exploration of different nodes for expansion, thereby enhancing the overall search efficiency.
\end{itemize}

\subsection{Comparison with Existing Methods}

In this section, we compare our proposed proof tree search method, which introduces a novel truncate-and-resume mechanism for whole-proof generation, with existing approaches. Current methods for using language models in formal mathematics proof search generally fall into two main strategies:
\begin{itemize}
    \item \textbf{Multi-pass proof-step generation}: This strategy breaks down the proving process into multiple episodes of tactic generation and verification, typically following a \textbf{tree search} pattern. It involves generating and verifying one tactic at a time, repeating the process for the next tactic until no proof goals remain. Notable examples include GPT-f~\citep{polu2020generative, polu2022formal}, Thor~\citep{jiang2022thor}, ReProver~\citep{yang2024leandojo}, Hypertree Proof Search~\citep{lample2022hypertree}, and InternLM2-StepProver~\citep{wu2024lean}.
    \item \textbf{Single-pass whole-proof generation}: This approach generates and verify an entire proof in one attempt. If the proof is incorrect, the model generates a new proof in the next attempt. Methods in this category include DSP~\citep{jiang2022draft}, Subgoal-Prover~\cite{zhao2023decomposing}, LEGO-Prover~\citep{wang2023lego}, Lyra~\citep{zheng2023lyra}, and miniCTX~\citep{hu2024minictx}.
\end{itemize}
Our proof tree search method uniquely bridges these two strategies, offering a novel hybrid approach. It starts with whole-proof generation, similar to the single-pass approach, but extends this by implementing a sophisticated truncate-and-resume mechanism. This process involves truncating the generated proof to its successful initial segment, parsing this segment into individual tactics, and resuming the tree search from this point. This iterative process effectively implements a Monte-Carlo Tree Search, seamlessly integrating single-pass whole-proof generation with multi-pass proof-step generation. Consequently, we can train a single model with nearly identical objectives to support both strategies simultaneously. Our experimental results demonstrate that this unified approach achieves superior performance in both settings. By combining the strengths of existing methods and introducing innovative techniques, our method offers a more versatile and effective solution for formal mathematics proof search, potentially paving the way for future advancements in this field.

\section{Experimental Results}
\label{sec:experiments}

In this section, we evaluate the theorem-proving capabilities of \thiswork~using two distinct benchmarks: miniF2F, which encompasses high-school level exercises and competition problems, and ProofNet, which pertains to undergraduate-level theorems. We present the results for both complete proof generation and Monte-Carlo tree search methodologies, utilizing the same trained model and inference configuration as Section~\ref{sec:sft-eval} to ensure consistency.

\subsection{Main Results}

We present a comparative analysis of \thiswork~against previous state-of-the-art language models, highlighting its performance and advancements.

\begin{itemize}
    \item \textbf{General-purpose Models:} \textbf{GPT-3.5} and \textbf{GPT-4}~\citep{OpenAI} are advanced generative AI models developed by OpenAI, known for their effectiveness across diverse tasks, including code generation. Despite not being specifically designed for theorem proving, their extensive parameter scales provide significant capabilities. The evaluation of these models in formal theorem proving is facilitated by \textbf{COPRA}~\citep{thakur2023language}, an in-context learning agent that leverages these large language models to propose tactic applications. Additionally, we examine \textbf{Llemma}~\citep{azerbayev2024llemma}, a series of language models trained on extensive general mathematical corpora, commonly used as the base model for formal theorem proving.

    \item \textbf{Specialized Models for Formal Mathematics:} \textbf{GPT-f}~\citep{polu2020generative, polu2022formal} represents an initial effort to apply Transformers~\citep{vaswani2017attention} to proof-step generation for theorem proving tasks, utilizing a best-first search module to construct complete proofs. Subsequent advancements include \textbf{ReProver}~\citep{yang2024leandojo}, \textbf{LLMStep}~\citep{welleck23llmstep}, and \textbf{Lean-STaR}~\citep{lin2024lean}. \textbf{Hypertree Proof Search}~\citep{lample2022hypertree} explores the use of Monte Carlo tree search in formal theorem proving using Lean. Concurrent works, \textbf{InternLM2-Math}~\citep{ying2024internlm} and \textbf{InternLM2-StepProver}~\citep{wu2024lean}, also demonstrate outstanding performance.
\end{itemize}

\paragraph{Metric.}
We compare the performance of \thiswork~with state-of-the-art models using the pass@$K$ accuracy metric, which evaluates the model's ability to generate a correct proof within $K$ attempts.
We display the sample budget $K$ according to the the following rules to align the computation budget across different generation schemes.
\begin{itemize}
    \item For single-pass sampling methods, we define the sample budget $K$ as the total number of proofs generated, with large values of $K$ factorized for the ease of comparison to tree search methods.
    \item For best-first-search methods, following the notation of \citet{azerbayev2024llemma}, we present $K=N\times S\times T$ where $N$ denotes the number of best-first-search attempts, $S$ denotes the number of tactics generated for each expansion, and $T$ denotes the number of expansion iterations.
    \item For tree search methods, \eg, \treesearch~and HTPS \citep{lample2022hypertree}, we present $K=N\times T$ where $N$ denotes the number of tree search attempts, and $T$ denotes the number of model generations invoked in tree expansions.
\end{itemize}

\paragraph{Results on miniF2F.}
Table~\ref{tab:minif2f_results} provides a comparative analysis of various theorem-proving methods on the miniF2F-test dataset. In the single-pass whole-proof generation setting, \thiswork-RL achieved the highest pass rate at 60.2\%, marking a significant improvement of 10.2 percentage points over \lastwork's 50.0\%. With a sampling budget limited to 128 attempts, \thiswork-RL proved 51.6\% of the problems, significantly outperforming other whole-proof generation methods and is comparable to the leading tree search methods. In the Tree Search Methods category, \thiswork-RL + \treesearch~ leads with a pass rate of 62.7\%, establishing a new state-of-the-art and creating a substantial gap with existing methods. Notably, \thiswork-RL requires only 3200 whole-proof generation samplings to achieve a pass rate of 54.9\%, surpassing the previous state-of-the-art result of InternLM2-StepProver, which performs $64\times 3200$ tree searches to achieve 54.5\%.

\begin{table*}[htp]
\setlength{\tabcolsep}{0.2in}
\begin{center}
\small
\begin{tabular}{lcc}
\toprule
    Method & Sample budget & miniF2F-test \\
    \toprule
    \multicolumn{3}{l}{\textit{Single-pass Whole-Proof Generation Methods}} \\
    \midrule
    TheoremLlama [\citenum{wang2024theoremllama}] & 128 & $33.6\%$ \\
    \midrule
    \multirow{2}{*}{\lastwork~[\citenum{xin2024deepseek}]} & $128$ & $46.1\%\pm 0.5\%$ \\
     & $16\times 4096$ & $50.0\%$ \\
    \midrule
    \multirow{3}{*}{\thiswork-Base} & 128 & $29.7\%\pm 0.5\%$ \\
    & 3200 & $39.2\%$ \\
    & 6400 & $42.2\%$ \\
    \midrule
    \multirow{6}{*}{\thiswork-SFT} & 32 & $48.2\% \pm 0.6\%$ \\
     & 64 & $49.6\% \pm 0.7\%$ \\
     & $128$ & $50.4\% \pm 0.4\%$ \\
     & $3200$ & $53.3\% \pm 0.5\%$ \\
     & $4\times 6400$ & $55.8\% \pm 0.7\%$ \\
     & $16\times 6400$ & $57.4\%$ \\
    \midrule
    \multirow{6}{*}{\thiswork-RL} & 32 & $50.0\%\pm 0.5\%$ \\
     & 64 & $50.7\%\pm 0.4\%$ \\
     & $128$ & $51.6\%\pm 0.5\%$ \\
     & $3200$ & $54.9\%\pm 0.7\%$ \\
     & $4\times 6400$ & $58.4\%\pm0.6\%$ \\
     & $16\times 6400$ & $\mathbf{60.2\%}$ \\
    \toprule
    \multicolumn{3}{l}{\textit{Tree Search Methods}} \\
    \midrule
    COPRA (Code Llama) [\citenum{thakur2023language}] & $1\times 500$ & $5.7\%$ \\
    COPRA (GPT-3.5) [\citenum{thakur2023language}] & $1\times 60$ & $9.0\%$ \\
    COPRA (GPT-4) [\citenum{thakur2023language}] & $1\times 60$ & $26.6\%$ \\
    Llemma-7B [\citenum{azerbayev2024llemma}] & $1\times 32\times 100$ & $26.2\%$ \\
    Llemma-34B [\citenum{azerbayev2024llemma}] & $1\times 32\times 100$ & $25.8\%$ \\
    ReProver [\citenum{yang2024leandojo}] & - & $26.5\%$ \\
    LLMStep [\citenum{welleck23llmstep}] & $1\times 32\times 100$ & $27.9\%$ \\
    GPT-f [\citenum{polu2022formal}] & $64\times 8\times 512$ & $36.6\%$\\
    Hypertree Proof Search [\citenum{lample2022hypertree}] & $64\times 5000$ & $41.0\%$ \\
    Lean-STaR [\citenum{lin2024lean}] & $64\times 1\times 50$ & $46.3\%$ \\
    InternLM2-Math-7B [\citenum{ying2024internlm}] & $1\times 32\times 100$ & $30.3\%$ \\
    InternLM2-Math-Plus-7B [\citenum{ying2024internlm}] & $1\times 32\times 100$ & $43.4\%$ \\
    \multirow{2}{*}{InternLM2-StepProver [\citenum{wu2024lean}]} & $1\times 32\times 100$ & $48.8\%$ \\
     & $64\times 32\times 100$ & $54.5\%$ \\
    \midrule
     \multirow{4}{*}{\thiswork-SFT + \treesearch} & $1\times 3200$ & $53.5\% \pm 0.4\%$ \\
     & $4\times 6400$ & $56.3\% \pm 0.3\%$ \\
     & $16\times 6400$ & $59.0\%$ \\
     & $32\times 6400^\dag$ & $60.2\%$ \\
    \midrule
    \multirow{4}{*}{\thiswork-RL + \treesearch} & $1\times 3200$ & $55.0\%\pm 0.7\%$ \\
     & $4\times 6400$ & $59.6\%\pm0.6\%$ \\
     & $16\times 6400$ & $\mathbf{62.7\%}$ \\
     & $32\times 6400^\dag$ & $\mathbf{63.5\%}$ \\
    \bottomrule
\end{tabular}
\caption{
Comparison with state-of-the-art methods on the miniF2F-test dataset. The notation $\mu\pm\sigma$ denotes the average accuracy $\mu$ and the standard deviation $\sigma$. Unless otherwise specified, \thiswork-Base results are based on 3-shot prompting, while \thiswork-SFT and RL employ CoT mode prompting. The symbol $\dag$ indicates performance using a mixture strategy with two guiding prompts (see Section~\ref{sec:large-train-eval} for details).
}
\label{tab:minif2f_results} 
\end{center}
\end{table*}

\paragraph{Results on ProofNet.}
Table~\ref{tab:proofnet_results} presents a comparative analysis of various theorem-proving methods on the ProofNet dataset. \thiswork-RL achieved pass rates of 22.6\% and 25.3\% for the overall ProofNet dataset in the single-pass whole-proof generation setting and with the enhancement of \treesearch, respectively. These results surpass the existing state-of-the-art methods, ReProver (13.8\%) and InternLM2-StepProver (18.1\%). When the number of whole-proof generation attempts is restricted to 3200, \thiswork~also proved 21.7\% of the theorems, demonstrating a 3.6\% improvement over the previous state-of-the-art, InternLM2-StepProver.

\begin{table*}[t!]
\setlength{\tabcolsep}{0.06in}
\begin{center}
\small
\begin{tabular}{lcccc}
\toprule
\multirow{2}{*}{Method} & \multirow{2}{*}{Sample budget} & \multicolumn{3}{c}{ProofNet} \\
 & & valid$^\ddag$ & test & all \\
\toprule
\multicolumn{4}{l}{\textit{Single-pass Whole-Proof Generation Methods}} \\
\midrule
\multirow{2}{*}{\thiswork-Base} & $128$ & $6.6\%\pm 0.9\%$ & $9.7\%\pm 0.7\%$ & $7.5\%\pm 0.7\%$ \\
& $3200$ & $10.8\%$ & $15.6\%$ & $13.2\%$  \\
\midrule
\multirow{3}{*}{\thiswork-SFT} & $128$ & $19.9\% \pm 0.4\%$ & $15.9\% \pm 0.6\%$ & $17.9\% \pm 0.3\%$ \\
& $3200$ & $20.7\% \pm 0.7\%$ & $21.0\% \pm 0.9\%$ & $20.9\% \pm 0.6\%$ \\
& $4\times 6400$ & $22.2\%$ & $23.7\%$ & $22.9\%$ \\
\midrule
\multirow{3}{*}{\thiswork-RL} & $128$ & $20.1\%\pm 0.5\%$ & $18.2\%\pm0.5\%$ & $19.1\%\pm 0.4\%$ \\
& $3200$ & $21.4\%\pm 0.3\%$ & $22.0\%\pm 0.5\%$ & $21.7\%\pm 0.4\%$ \\
& $4\times 6400$ & $21.6\%$ & $23.7\%$ & $22.6\%$ \\
\toprule
\multicolumn{4}{l}{\textit{Tree Search Methods}} \\
\midrule
ReProver [\citenum{yang2024leandojo}] & - & - & - & $13.8\%$ \\
InternLM2-StepProver [\citenum{wu2024lean}] & $1\times 32\times 100$ & - & - & $18.1\%$ \\
\midrule
\multirow{2}{*}{\thiswork-SFT + \treesearch} & $1\times 3200$ & $22.2\% \pm 0.7\%$ & $21.6\% \pm 0.2\%$ & $21.9\% \pm 0.4\%$ \\
 & $4\times 6400$ & $23.8\%$ & $\textbf{25.8\%}$ & $24.8\%$ \\
\midrule
\multirow{2}{*}{\thiswork-RL + \treesearch} & $1\times 3200$ & $22.0\%\pm 0.3\%$ & $21.5\%\pm0.8\%$ & $21.8\%\pm0.4\%$ \\
 & $4\times 6400$ & $25.4\%$ & $\textbf{25.3\%}$ & $25.3\%$ \\
\bottomrule
\end{tabular}
\caption{\centering
Comparing with state-of-the-arts on the ProofNet dataset. $^\ddag$ Note that the validation set of ProofNet is used to perform expert iteration in supervised fine-tuning.
}
\label{tab:proofnet_results} 
\end{center}
\end{table*}

\vspace{-0.2in}

\subsection{Re-Examining the Effectiveness of Training Strategies on Large-scale Sampling} \label{sec:large-train-eval}

We revisit the effects of several training modules in n a large-scale sampling setting, focusing on both single-pass whole-proof generation and Monte-Carlo tree search. The results demonstration that the observations and findings presented in Section~\ref{sec:sft-eval} generalize to sampling scenarios with a large sample size.

\vspace{-0.1in}

\paragraph{General Enhancement of Reinforcement Learning.}
To support the claim that online reinforcement learning from verification feedback generally enhances the model capabilities, we compare our final model to the SFT-only version using a large sample budget.
The comparison results are presented as two columns in Table~\ref{tab:training-ablation}.
\thiswork-RL consistently outperforms the SFT model across all generation settings, regardless of whether the chain-of-thought strategy is applied.
The results also indicate that the improvements gained from conducting online RL is orthogonal to those achieved through \treesearch, which can be further combined to boost the performance.
By integrating both CoT prompting and \treesearch, \thiswork-RL achieves a pass rate of $62.7\%$ on miniF2F-test.
This performance shows a notable $3.7\%$ improvement over the SFT model, highlighting the critical role of reinforcement learning in enhancing the overall effectiveness of the proof completion model.

\vspace{-0.1in}

\paragraph{CoT, non-CoT, and Mixture Strategy.}
We compare the performance of two generation modes, \ie, non-CoT and CoT, on miniF2F-test dataset.
The results, shown in Table~\ref{tab:training-ablation}, indicate that the advantage of CoT over the non-CoT mode is amplified as the sample budget increases.
This suggests that the incorporation of natural language chain-of-thought can diversify the planning pathways of theorem proving, potentially leading to a broader range of reasoning strategies and more innovative solutions.
Results also show that these two modes have complementary advantages across different problems.
The model's theorem proving strategy in the CoT mode is more systematic and proactive in mathematical thinking, while in the non-CoT mode, the model can efficiently use Lean high-level tactics to solve computational problems that can be addressed within Lean’s automation mechanisms.
To leverage these advantages, we consider a mixture strategy, denoted by non-CoT \& CoT in Table~\ref{tab:training-ablation}, allocates half of sample budget to the CoT mode and the remains to the non-CoT mode.
This simple combination of two guiding prompts shows great promise in further bootstrapping the performance of our proof completion model, achieving a pass rate of $63.5\%$ on miniF2F-test. In Appendix~\ref{apx:solutions}, we present example problems that illustrate the different advantages of the two generation modes.

\begin{table*}[t!]
\setlength{\tabcolsep}{0.17in}
\begin{center}
\small

\begin{tabular}{ccccc}
\toprule \vspace{0.02in}
    & \multirow{2}{*}{Prompt mode} & \multirow{2}{*}{Sample budget} & \multicolumn{2}{c}{\thiswork} \\
    & & & SFT & RL \\
    \toprule
    & \multirow{2}{*}{non-CoT} & $4\times 6400$ & $54.7\%\pm0.4\%$ & $56.5\%\pm 0.5\%$ \\
    & & $16\times 6400$ & $56.1\%$ & $57.4\%$ \\ \cline{2-5}
    \multirow{2}{*}{Single-Pass} & \multirow{2}{*}{CoT} & $4\times 6400$ & $55.8\%\pm 0.7\%$ & $58.4\%\pm0.5\%$ \\
    \multirow{2}{*}{Generation} & & $16\times 6400$ & $57.4\%$ & $60.2\%$ \\ \cline{2-5}
    & \multirow{3}{*}{non-CoT \& CoT} & $(2+2)\times 6400$ & $56.1\%\pm0.8\%$ & $58.3\%\pm0.6\%$ \\
    & & $(8+8)\times 6400$ & $58.2\%$ & $60.7\%$ \\
    & & $(16+16)\times 6400$ & $58.6\%$ & $61.1\%$ \\
    \midrule
    \multirow{7}{*}{\treesearch} & \multirow{2}{*}{non-CoT} & $4\times 6400$ & $55.7\%\pm 0.6\%$ & $58.4\%\pm 0.6\%$ \\
    & & $16\times 6400$ & $57.8\%$ & $59.4\%$ \\ \cline{2-5}
    & \multirow{2}{*}{CoT} & $4\times 6400$ & $56.3\%\pm 0.3\%$ & $59.6\%\pm 0.6\%$ \\
    & & $16\times 6400$ & $59.0\%$ & $62.7\%$ \\ \cline{2-5}
    & \multirow{3}{*}{non-CoT \& CoT} & $(2+2)\times 6400$ & $56.1\%\pm 0.8\%$ & $60.0\%\pm 0.8\%$ \\
    & & $(8+8)\times 6400$ & $59.0\%$ & $63.1\%$ \\
    & & $(16+16)\times 6400$ & $60.2\%$ & $\textbf{63.5\%}$ \\
    \bottomrule
\end{tabular}
\caption{\centering
A large-scale ablation study to investigate the effectiveness of several algorithmic designs on model training. The results are evaluated on the miniF2F-test dataset.
}
\label{tab:training-ablation} 
\end{center}
\end{table*}

\subsection{Ablation Studies on \treesearch} \label{sec:rmaxts-ablation}

\begin{figure}[t!]
\setlength{\tabcolsep}{0.09in}
\begin{minipage}{0.48\textwidth}
    \centering
    \includegraphics[width=0.98\textwidth]{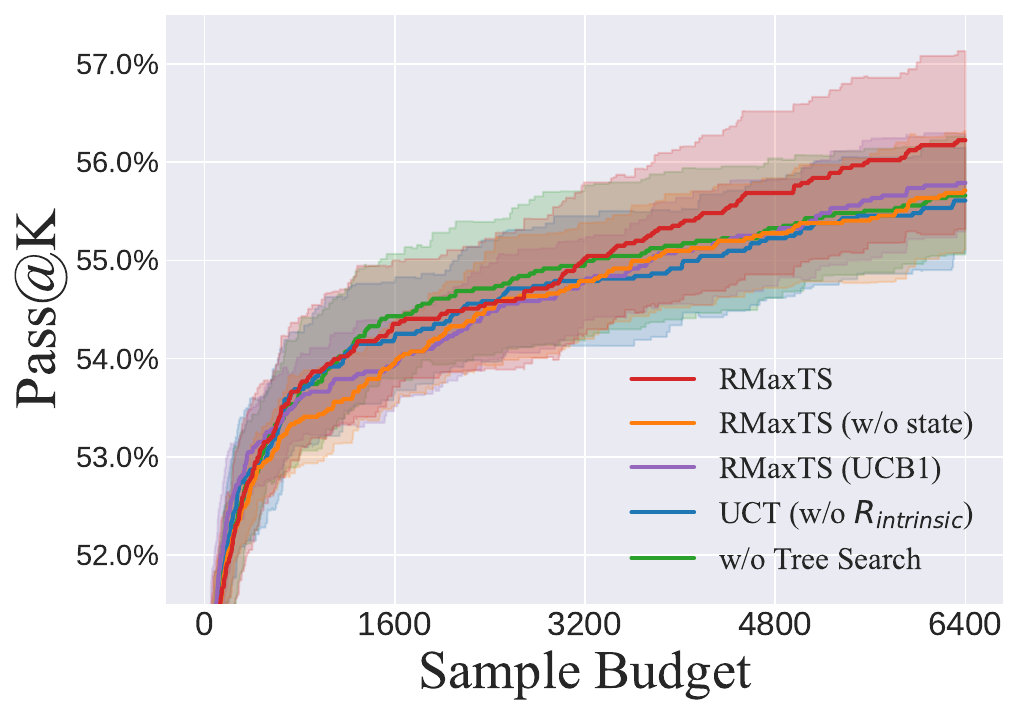}
\end{minipage}
\begin{minipage}{0.52\textwidth}
\begin{center}
    \scriptsize \vspace{-0.2in}
    \begin{tabular}{ccc}
        \toprule \vspace{0.02in}
        & Sample budget & miniF2F-test \\
        \toprule
        \multirow{2}{*}{Single-Pass Generation} & $4\times 6400$ & $58.4\% \pm 0.5\%$ \\
        & $16\times 6400$ & $60.2\%$ \\
        \midrule
        UCT & $4\times 6400$ & $58.2\% \pm 0.3\%$ \\
        (without $R_{\text{intrinsic}}$) & $16\times 6400$ & $61.1\%$ \\
        \midrule
        \treesearch & $4\times 6400$ & $58.6\% \pm 0.3\%$ \\
        (DUCB $\rightarrow$ UCB1) & $16\times 6400$ & $60.7\%$ \\
        \midrule
        \treesearch & $4\times 6400$ & $58.4\% \pm 0.3\%$ \\
        (without tactic state) & $16\times 6400$ & $61.1\%$ \\
        \midrule
        \multirow{2}{*}{\treesearch} & $4\times 6400$ & $59.6\% \pm 0.6\%$ \\
        & $16\times 6400$ & $62.7\%$ \\
        \bottomrule
    \end{tabular}
\end{center}
\end{minipage}
\caption{A modular ablation study examining the algorithmic design of \treesearch. The experiments are conducted on the miniF2F-test dataset with \thiswork-RL using the CoT mode. The left panel presents the curves of Pass@K accuracy within 6400 generation samples. The results with a larger sample size are presented in the right panel.}
\label{fig:rmaxts-ablation} 
\end{figure}

\paragraph{Intrinsic Rewards and Discounted UCB.}
We investigate the effectiveness of two core components of \treesearch, \ie, the intrinsic rewards defined in Eq.~\eqref{eq:rmax_reward} and the discounted upper confidence bound stated in Eq.~\eqref{eq:ducb}.
We start with a baseline implementing the standard UCT algorithm \citep{kocsis2006bandit} without intrinsic rewards, in which the exploration is driven exclusively by the UCB bonus.
Note that, since no non-zero rewards are provided for this baseline, all variants of the UCB formula become equivalent, as node selection is determined solely by visitation counts.
The experimental results in Figure~\ref{fig:rmaxts-ablation} show that, in the absence of intrinsic rewards, the performance of UCT (without $R_{\text{intrinsic}}$) degenerates into a level comparable to that of non-search methods.
Furthermore, we consider \treesearch~using the standard UCB1 (refer to Eq.~\eqref{eq:ucb1}) instead of the discounted UCB, denoted by \treesearch~(DUCB $\rightarrow$ UCB1).
The results indicate that the performance of \treesearch~with UCB1 bonus is also moderate, comparable to that of UCT (without $R_{\text{intrinsic}}$).
That is because UCB1 is designed to guarantee asymptotic performance through exhausted exploration \citep{auer2002finite} assuming the sample size to be sufficiently large.
In contrast, the discounted UCB can accelerate the value propagation of non-stationary intrinsic rewards, preventing the guidance of $R_{\text{intrinsic}}$ from being dominated by that of visitation counts.
This demonstrates that the discounted UCB mechanism is a crucial complement to intrinsic-reward-driven exploration.

\paragraph{Guidance of Tactic State Information.}
When expanding a tree node, we concatenate the intermediate tactic state information as a comment block to the incomplete code to guide the proof completion.
With the provided auxiliary information, the proof completion model can enhance its internal representation of the tactic state, offering intermediate guidance for long-horizon planning.
To demonstrate this advantage, we present experiments on \treesearch~that performs code completion directly from the raw incomplete code without accessing tactic state information, denoted by \treesearch~(without tactic state) in Figure~\ref{fig:rmaxts-ablation}.
The results indicate that the performance gain from applying tree search becomes moderate in the absence of tactic state information, especially when tackling hard problems that require a large amount of samples.
This highlights that the integration of compiler information is an essential component of the tree search algorithm, enhancing its overall effectiveness and sample efficiency.

\section{Conclusion, Limitation, and Future Work}

We present \thiswork, a language model with 7 billion parameters that outperforms all open-source models in formal theorem proving in Lean 4.
\thiswork~is initialized with \thiswork-Base, which extends the pre-training of DeepSeekMath-Base 7B using a specialized corpus for formal mathematical reasoning.
Supervised fine-tuning is conducted on a comprehensive Lean 4 code completion dataset, encompassing a wide range of formal theorems from various mathematical domains.
Subsequently, we employ GRPO to enhance whole-proof generation through online reinforcement learning.
Upon developing the \thiswork~model, we introduce \treesearch, a variant of Monte-Carlo tree search, to improve problem-solving capabilities via large-scale search with extensive exploration.
These components form a comprehensive pipeline for training an LLM-based proof assistant, enabling \thiswork~to achieve significant improvements over \lastwork.

The framework of \thiswork~is designed to establish an AlphaZero-like pipeline for formal theorem proving.
The use of expert iteration and synthetic data mirrors the core trial-and-error loop of reinforcement learning, with the compiler oracle serving as the world model to provide environmental supervision.
Within the RL paradigm, the integrated tree search module has proven to be highly effective in advancing superhuman performance across various domains \citep{silver2016mastering, fawzi2022discovering, lutz2023top}.
In developing \thiswork, we focus on the exploration aspect of RL, introducing \treesearch~to diversify the generation of proof steps.
However, the exploitation aspect, specifically the problem of proof search, remains unexplored.
A promising future direction is training a critic model to assess incomplete proofs and prune search branches.
Such a partial-proof critic model would implicitly perform temporal credit assignment \citep{sutton1984temporal}, decomposing proof-level feedback into step-wise value differences \citep{arjona2019rudder}.
Developing critic models for assessing long planning paths and providing guidance rewards presents a crucial and challenging problem \citep{ng2000algorithms, sorg2010reward} that warrants further investigation.

Finally, recent work has progressed beyond proving individual theorems to addressing real-world theory proving within complex, multi-theorem Lean files~\citep{hu2024minictx}. This shift is a natural extension of our whole-proof generation approach. Our observations indicate that the current model already possesses some understanding of file-level context. Moving forward, we will focus on enhancing this aspect to support cutting-edge Lean mathematical formalization developers with our language model advancements.

\setcitestyle{numbers}
\bibliographystyle{abbrvnat}
\bibliography{main}
\setcitestyle{authoryear}

\newpage
\appendix

\section{\large Illustrative Examples of Non-CoT and CoT Prompting for Proof Completion}
\label{app:promt_examples}

In this section, we present examples of non-CoT and CoT prompting and the generated proofs for $\mathtt{aime\_1983\_p9}$, a problem from the miniF2F benchmark~\citep{zheng2021minif2f}. For clarity, inline LaTeX expressions are utilized. Firstly, here is an example of whole-proof generation in non-CoT prompting mode:

\noindent \textit{Input}:
\begin{lstlisting}[frame=single]
Complete the following Lean 4 code:

```lean4
import Mathlib
import Aesop

set_option maxHeartbeats 0

open BigOperators Real Nat Topology Rat

/-- Show that $\frac{9x^2\sin^2 x + 4}{x\sin x} \geq 12$ for $0 < x < \pi$.-/
theorem aime_1983_p9 (x : ℝ) (h₀ : 0 < x ∧ x < Real.pi) :
  12 ≤ (9 * (x ^ 2 * Real.sin x ^ 2) + 4) / (x * Real.sin x) := by
\end{lstlisting}
\noindent \textit{Output}:
\begin{lstlisting}[frame=single]
  rw [le_div_iff (mul_pos h₀.1 (Real.sin_pos_of_pos_of_lt_pi h₀.1 h₀.2))]
  nlinarith [mul_self_nonneg (3 * x * Real.sin x - 2)]
```
\end{lstlisting}

Next, here is an example of whole-proof generation in CoT prompting mode. This proof was found using a Monte-Carlo tree search with a truncate-and-resume mechanism (see Section~\ref{sec:tree-search} for details). The model receives a prompt including a proof code in progress, ending with a comment block containing the intermediate tactic state from the Lean 4 prover. The subsequent code is generated with guidance from this annotated code prefix.

\noindent \textit{Input}:
\begin{lstlisting}[frame=single]
Complete the following Lean 4 code with explanatory comments preceding each line of code:

```lean4
import Mathlib
import Aesop

set_option maxHeartbeats 0

open BigOperators Real Nat Topology Rat

/-- Show that $\frac{9x^2\sin^2 x + 4}{x\sin x} \geq 12$ for $0 < x < \pi$.-/
theorem aime_1983_p9 (x : ℝ) (h₀ : 0 < x ∧ x < Real.pi) :
  12 ≤ (9 * (x ^ 2 * Real.sin x ^ 2) + 4) / (x * Real.sin x) := by
  /-
  To find the minimum value of $\frac{9x^2\sin^2 x + 4}{x\sin x}$ for $0 < x < \pi$, we need to show that it is at least 12. We start by noting that the expression can be rewritten using the division property of inequalities. We then use the fact that $\sin x$ and $x$ are positive in the given range to establish the necessary inequalities. Finally, we apply these results to conclude that the minimum value is indeed 12.
  -/
  -- We start by ensuring that the product $x * \sin x$ is positive in the given range.
  have h₁ : 0 < x * Real.sin x := by
    apply mul_pos
    -- $x$ is positive in the range $(0, \pi)$.
    exact h₀.1
    -- $\sin x$ is positive in the range $(0, \pi)$.
    exact Real.sin_pos_of_pos_of_lt_pi h₀.1 h₀.2
  -- Using the division property of inequalities, we rewrite the expression.
  rw [le_div_iff h₁]
  /- tactic state:
    x : ℝ
    h₀ : 0 < x ∧ x < π
    h₁ : 0 < x * x.sin
    ⊢ 12 * (x * x.sin) ≤ 9 * (x ^ 2 * x.sin ^ 2) + 4
  -/
\end{lstlisting}
\noindent \textit{Output}:
\begin{lstlisting}[frame=single]
  -- This is equivalent to showing that $9x^2 \sin^2 x - 12x \sin x + 4 \geq 0$, and the left-hand side can be rewritten as a perfect square $(3x \sin x - 2)^2$.
  -- We use the fact that $(3x \sin x - 2)^2$ is non-negative to establish this.
  nlinarith [sq_nonneg (3 * x * Real.sin x - 2)]
```
\end{lstlisting}

In this case, the proof completion model first extracts the goal expression from the tactic state, \ie, the inequality $12(x\sin x)\leq 9(x^2\sin^2 x) + 4$ at the last line of the comment block, and then arranges it into the form $9x^2 \sin^2 x - 12x \sin x + 4 \geq 0$ as part of the chain-of-thought process. Finally, the left-hand side is rewritten as a square term to establish non-negativity, closing the proof.

\section{Example Solutions to MiniF2F-test Problems} \label{apx:solutions}

To illustrate the different advantages and complementary of non-CoT mode and CoT mode of prompting, we provide examples where non-CoT mode succeeds while CoT mode fails, and vice versa.

\subsection{Problems that non-CoT outperforms CoT}

\begin{lstlisting}[frame=single]
theorem induction_pord1p1on2powklt5on2 (n : ℕ) (h₀ : 0 < n) :
    ($\Pi$ k in Finset.Icc 1 n, 1 + (1 : ℝ) / 2 ^ k) < 5 / 2 := by
  rw [Finset.Icc]
  simp_all [Nat.succ_le_iff, Nat.one_le_iff_ne_zero]
  have h₁ : 0 < 2 ^ k := by apply pow_pos <;> norm_num
  norm_num
  have h₂ : (2 : ℝ) ≠ 0 := by norm_num
  field_simp
  rw [div_lt_div_iff]
  ring_nf
  norm_cast
  nlinarith
  all_goals norm_cast
  all_goals linarith
\end{lstlisting}

\begin{lstlisting}[frame=single]
theorem imo_1960_p2 (x : ℝ) (h₀ : 0 ≤ 1 + 2 * x) (h₁ : (1 - Real.sqrt (1 + 2 * x)) ^ 2 ≠ 0)
    (h₂ : 4 * x ^ 2 / (1 - Real.sqrt (1 + 2 * x)) ^ 2 < 2 * x + 9) : -(1 / 2) ≤ x ∧ x < 45 / 8 := by
  norm_num at h₀ h₁ h₂
  have h₃ : 0 ≤ 1 + 2 * x := by linarith
  have h₄ : 0 < 1 + Real.sqrt (1 + 2 * x) := by
    nlinarith [Real.sqrt_nonneg (1 + 2 * x)]
  have h₅ : 4 * x ^ 2 / (1 - Real.sqrt (1 + 2 * x)) ^ 2 < 2 * x + 9 := by linarith
  have h₆ : 1 - Real.sqrt (1 + 2 * x) ≠ 0 := by
    intro h
    apply h₁
    nlinarith
  have h₇ : 4 * x ^ 2 / (1 - Real.sqrt (1 + 2 * x)) ^ 2 = (1 + Real.sqrt (1 + 2 * x)) ^ 2 := by
    field_simp [h₆]
    nlinarith [sq_sqrt (show 0 ≤ 1 + 2 * x by linarith)]
  rw [h₇] at h₅
  constructor <;> nlinarith [sq_sqrt (show 0 ≤ 1 + 2 * x by linarith)]
\end{lstlisting}

\subsection{Problems that CoT outperforms non-CoT}

\begin{lstlisting}[frame=single]
theorem mathd_algebra_459 (a b c d : ℚ) (h₀ : 3 * a = b + c + d) (h₁ : 4 * b = a + c + d)
    (h₂ : 2 * c = a + b + d) (h₃ : 8 * a + 10 * b + 6 * c = 24) : ↑d.den + d.num = 28 := by
  /-
  Given the system of equations involving the times that Anna (a), Bertram (b), Carli (c), and David (d) can hold their breaths:
  1. $ 3a = b + c + d $
  2. $ 4b = a + c + d $
  3. $ 2c = a + b + d $
  4. $ 8a + 10b + 6c = 24 $
  We need to find the sum of the numerator and the denominator of the fraction representing David's time in minutes, simplified to its lowest terms, and show that it is 28.
  First, we solve the system of equations step by step:
  - From the first equation, we can express $ d $ in terms of $ a, b, $ and $ c $:
    $ d = 3a - b - c $
  - Substitute $ d $ into the second equation:
    $ 4b = a + c + (3a - b - c) $
    Simplifying, we get:
    $ 4b = 4a - b $
    $ 5b = 4a $
    $ b = \frac{4a}{5} $
  - Substitute $ b $ into the third equation:
    $ 2c = a + \frac{4a}{5} + (3a - b - c) $
    Simplifying, we get:
    $ 2c = a + \frac{4a}{5} + 3a - \frac{4a}{5} - c $
    $ 2c = 4a - c $
    $ 3c = 4a $
    $ c = \frac{4a}{3} $
  - Substitute $ b $ and $ c $ into the fourth equation:
    $ 8a + 10 \left(\frac{4a}{5}\right) + 6 \left(\frac{4a}{3}\right) = 24 $
    Simplifying, we get:
    $ 8a + 8a + 8a = 24 $
    $ 24a = 24 $
    $ a = 1 $
  Now, we find $ b $ and $ c $:
  - $ b = \frac{4a}{5} = \frac{4}{5} $
  - $ c = \frac{4a}{3} = \frac{4}{3} $
  Finally, we find $ d $:
  - $ d = 3a - b - c = 3 \cdot 1 - \frac{4}{5} - \frac{4}{3} = 3 - \frac{4}{5} - \frac{4}{3} = \frac{45}{15} - \frac{12}{15} - \frac{20}{15} = \frac{13}{15} $
  The fraction $ \frac{13}{15} $ is already in its simplest form, and the sum of the numerator and the denominator is:
  $ 13 + 15 = 28 $
  -/
  have h₄ : a = 1 := by
    have h₅ : 8 * a + 10 * b + 6 * c = 24 := h₃
    have h₆ : 24 * a = 24 := by
      -- Solve the system of equations to find the value of a
      linarith
    exact mul_right_cancel₀ (show (24 : ℚ) ≠ 0 by norm_num) 
      (by linarith)
  have h₅ : b = 4 / 5 := by
    -- Solve for b using the value of a
    linarith
  have h₆ : c = 4 / 3 := by
    -- Solve for c using the value of a
    linarith
  have h₇ : d = 13 / 15 := by
    -- Solve for d using the values of a, b, and c
    linarith
  -- Sum the numerator and the denominator of d
  norm_num [h₇]
  <;> rfl
\end{lstlisting}

\begin{lstlisting}[frame=single]
theorem numbertheory_x5neqy2p4 (x y : ℤ) : x ^ 5 ≠ y ^ 2 + 4 := by
  /-
  To show that for any two integers $ x $ and $ y $, $ x^5 \neq y^2 + 4 $, we can use modular arithmetic to demonstrate that the equation $ x^5 \equiv y^2 + 4 \pmod{11} $ has no solutions. Specifically, we will consider the possible values of $ x $ and $ y $ modulo 11 and show that none of these combinations satisfy the equation.
  -/
  -- Use modular arithmetic to show that the equation has no solutions modulo 11.
  have : ¬(x ^ 5 % 11 = (y ^ 2 + 4) % 11) := by
    -- Consider all possible values of x modulo 11.
    have : x % 11 = 0 ∨ x % 11 = 1 ∨ x % 11 = 2 ∨ x % 11 = 3 ∨ x % 11 = 4 ∨ x % 11 = 5 ∨
        x % 11 = 6 ∨ x % 11 = 7 ∨ x % 11 = 8 ∨ x % 11 = 9 ∨ x % 11 = 10 := by
      omega
    -- Consider all possible values of y modulo 11.
    have : y % 11 = 0 ∨ y % 11 = 1 ∨ y % 11 = 2 ∨ y % 11 = 3 ∨ y % 11 = 4 ∨ y % 11 = 5 ∨
        y % 11 = 6 ∨ y % 11 = 7 ∨ y % 11 = 8 ∨ y % 11 = 9 ∨ y % 11 = 10 := by
      omega
    -- For each combination of x and y modulo 11, check if the equation holds.
    rcases this with (h | h | h | h | h | h | h | h | h | h | h) <;>
      rcases this with (h' | h' | h' | h' | h' | h' | h' | h' | h' | h' | h') <;>
        simp [h, h', pow_succ, Int.mul_emod, Int.add_emod]
  -- Use the result from modular arithmetic to conclude the theorem.
  omega
\end{lstlisting}

\begin{lstlisting}[frame=single]
theorem amc12_2000_p12 (a m c : ℕ) (h₀ : a + m + c = 12) :
    a * m * c + a * m + m * c + a * c ≤ 112 := by
  /-
  Given nonnegative integers $A, M, $ and $ C $ such that $ A + M + C = 12 $, we need to find the maximum value of $ A \cdot M \cdot C + A \cdot M + M \cdot C + A \cdot C $.
  To solve this, we can use the fact that the expression $ A \cdot M \cdot C + A \cdot M + M \cdot C + A \cdot C $ is a polynomial in terms of $ A, M, $ and $ C $. We can try different values of $ A, M, $ and $ C $ that sum to 12 and compute the value of the polynomial for each combination. The maximum value of the polynomial will then be the solution.
  -/
  -- We use a lemma that states if the sum of three numbers is less than or equal to a certain value, then the product of these numbers is also less than or equal to a certain value.
  have h₁ : a ≤ 12 ∧ m ≤ 12 ∧ c ≤ 12 := by
    constructor
    · -- Since a + m + c = 12, a cannot be greater than 12.
      omega
    · constructor
      · -- Similarly, m cannot be greater than 12.
        omega
      · -- And c cannot be greater than 12.
        omega
  -- Using the lemma, we compute the value of the polynomial for different values of a, m, and c that sum to 12.
  cases' h₁ with h₁_left h₁_right
  cases' h₁_right with h₁_middle h₁_right
  -- We use interval_cases to try different values of a, m, and c.
  interval_cases a <;> interval_cases m <;> interval_cases c <;>
    -- For each combination, we compute the value of the polynomial and check if it is less than or equal to 112.
    simp_all (config := {decide := true})
\end{lstlisting}

\end{CJK*}
\end{document}